\documentclass{article}

\usepackage[preprint]{neurips_2026}

\usepackage[utf8]{inputenc}
\usepackage[T1]{fontenc}
\usepackage{microtype}
\usepackage{graphicx}
\usepackage{subcaption}
\usepackage{booktabs}
\usepackage[pagebackref,breaklinks,colorlinks,allcolors=citeblue]{hyperref}
\usepackage{url}

\usepackage{amsmath}
\usepackage{amssymb}
\usepackage{mathtools}
\usepackage{amsthm}
\usepackage{amsfonts}
\usepackage{multirow}
\usepackage{makecell}
\usepackage{colortbl}
\usepackage[table]{xcolor}
\usepackage{threeparttable}
\usepackage{algorithm}
\usepackage{algorithmic}

\newcommand{\ql}[1]{\textcolor{black}{#1}}
\newcommand{\yh}[1]{\textcolor{black}{#1}}

\usepackage[capitalize,noabbrev]{cleveref}

\theoremstyle{plain}

\theoremstyle{definition}

\theoremstyle{remark}

\definecolor{citeblue}{rgb}{0.21,0.49,0.74}
\usepackage[textsize=tiny]{todonotes}

\title{Geometry and Gradient-based Partitioning for Panoramic Outdoor Reconstruction}

\author{
  Weijian Chen$^{1,2}$, Weibo Yao$^{1,3}$, Yuhang Zhang$^1$, Xiaolin Tang$^1$, Guo Wang$^1$,\\\textbf{Weijun Zhang$^1$, 
  Xitong Gao$^4$, Yihao Chen$^1$, Hongde Qin$^5$, 
  Lu Qi$^{1,6}$\thanks{Corresponding author: Lu Qi}} \\
  $^1$Insta360 Research, $^2$ Sun Yat-sen University, $^3$ South China University of Technology, \\
  $^4$ University of Chinese Academy of Sciences, 
  $^5$ Harbin Engineering University, 
  $^6$ Wuhan University\\
}

\begin{document}

\maketitle

\begin{figure}[!h]
    \centering
    \includegraphics[width=\linewidth]{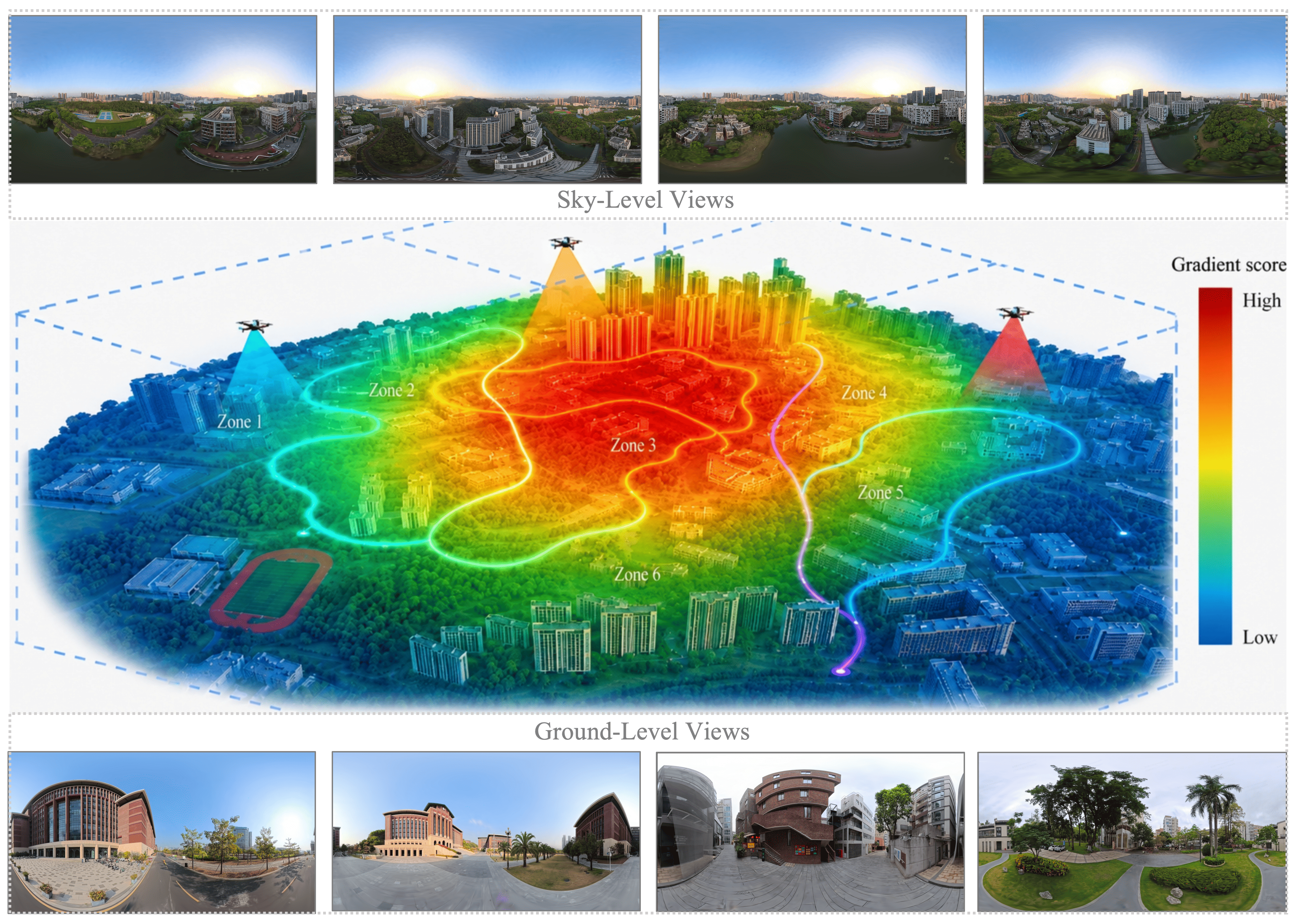}
    \caption{Overview of PanoLOG, which enables scalable large-scale outdoor reconstruction by introducing G$^2$PS, a Geometry and Gradient-based Partitioning Strategy for $360^{\circ}$ panoramic scenes.
}
    \label{fig: teaser}
\end{figure}

\begin{abstract}
Scaling 3D Gaussian Splatting (3DGS) to large outdoor scenes is costly in both data acquisition and computation. 
\ql{Adopting} panoramic images with equirectangular projection (ERP) \ql{can} reduce capture effort via their full $360^{\circ}$ field of view, yet the resulting omnipresent visibility invalidates existing partitioning strategies that rely on local camera frustums, causing block-wise optimization to degenerate into global training.
\ql{Thus}, we propose PanoLOG, a two-stage coarse-to-fine framework equipped with a Geometry and Gradient-based Partitioning Strategy (G$^2$PS) tailored for large-scale panoramic 3DGS reconstruction.
%
% In the global coarse stage, PanoLOG incorporates explicit sky sphere modeling and panoramic monocular depth supervision to establish reliable scene geometry; in the block-wise refinement stage, G$^2$PS constructs an adaptive bounding volume via parallax-driven depth uncertainty analysis and employs gradient-based importance scoring to allocate cameras to spatial blocks according to their actual observational contributions.
\ql{In the global coarse stage, PanoLOG leverages sky-sphere modeling and panoramic monocular depth supervision for reliable geometry, while in the refinement stage, G$^2$PS builds adaptive bounding volumes via parallax-driven uncertainty and assigns cameras via gradient-based importance scoring.}
Furthermore, we construct Pano360, the first benchmark on large-scale panoramic dataset for outdoor scene reconstruction.
Extensive experiments demonstrate that G$^2$PS achieves state-of-the-art rendering quality while maintaining scalable, block-parallel training.
\ql{Our models, training code, and dataset are publicly available \href{https://insta360-research-team.github.io/GGPS-Website/}{https://insta360-research-team.github.io/GGPS-Website/}.}

\end{abstract}

\section{Introduction}
% Novel View Synthesis~\citep{chung2023luciddreamer, wu2025unigs,wu20244d} is a popular research area and benefits widespread applications ranging from autonomous driving simulation to virtual reality (VR).
\ql{Novel View Synthesis~\citep{chung2023luciddreamer, wu2025unigs, wu20244d} has emerged as a cornerstone of computer vision, underpinning applications from autonomous driving simulation to virtual reality. Recently, 3D Gaussian Splatting (3DGS) has become the mainstream paradigm for photorealistic reconstruction, leveraging anisotropic explicit Gaussian primitives and an efficient differentiable rasterization pipeline to achieve unprecedented rendering speed.}

\ql{Despite its success, scaling 3DGS to expansive outdoor environments remains challenging due to the prohibitive costs of data acquisition and computational overhead. Conventional pipelines typically rely on capturing a massive volume of narrow field-of-view (FoV) images, subsequently adopting a divide-and-conquer strategy to partition the scene. However, such workflows are not only labor-intensive but also prone to structural inconsistencies across stitching boundaries, limiting their efficiency for large-scale deployment.}

\ql{Inspired by omnidirectional vision, adopting a panoramic ERP representation~\citep{lin2025one,ge2026airsim360,feng2026dit360} offers a compelling alternative by capturing a full $360^{\circ}$ field of view in a single shot, thereby streamlining data collection. 
Yet, this shift moves the complexity from acquisition to spatial partitioning. 
\yh{Traditional partitioning ~\citep{kerbl2024hierarchical, liu2024citygaussian, lin2024vastgaussian} relies on the local visibility of pinhole cameras, which is at odds with the $360^{\circ}$ nature of panoramic vision. 
Without frustum constraints, the omnipresent visibility in panoramic scenes strips these strategies of their discriminative power, causing intended block-wise partitioning to degenerate back into inefficient global optimization.}
}
\ql{Thus, one question raised: could we have a specific partition strategy that can adapt for panoramic images?}

\ql{To address this issue, we propose G$^2$PS, a two-stage coarse-to-fine training pipeline alongside a Geometry and Gradient-based Partitioning Strategy (G$^2$PS).
%
% Specifically, \wj{G$^2$PS is guided by panoramic geometric constraints and gradient distributions.
%
In the global coarse stage, we estimate a relative parallax distance using inter-camera baselines and perform depth uncertainty analysis, enabling adaptive expansion of spatial boundaries to convert unbounded scenes into stable reconstruction regions.
In the local refinement stage, we leverage gradient signals from the coarse stage to evaluate each camera's actual contribution to different spatial blocks, enabling gradient-based importance assignment that effectively resolves the camera--block allocation problem in panoramic scenes.
%
% Furthermore, to address the pervasive challenges of low-texture regions and unbounded sky in outdoor scenes, we design a dedicated optimization framework. By incorporating explicit sky sphere modeling and panoramic monocular depth estimation, we provide stable depth supervision in regions lacking geometric constraints, reducing artifacts and improving overall reconstruction quality.
}
\yh{In this way, we obtain structurally sound spatial partitioning, enabling each sub-block to focus on its key observation regions, thereby enhancing the quality of block optimization and preventing global degeneracy.}
% }

\ql{To well evaluate our performance, we build a comprehensive benchmark that collects outdoor images \yh{across four panoramic scenes (campus and park via Antigravity A1; two street scenes via Insta360 X5). Covering over 2 million $m^2$ with 5,637 images (3840$\times$1920), we construct a dataset named \textbf{Pano360}, which can provides calibrated poses and sparse point clouds for reproducible evaluation.}}

\ql{Our contributions are threefold.}
\begin{itemize}
    \item We present PanoLOG, a novel 3DGS framework for large-scale outdoor panoramic reconstruction using a two-stage coarse-to-fine pipeline. By shifting from inefficient global optimization to a block-wise paradigm, PanoLOG effectively addresses visibility challenges in panoramic image, enabling high-fidelity, scalable scene modeling.
    \item We introduce the Geometry and Gradient-based Partitioning Strategy (G$^2$PS). It leverages parallax-driven depth uncertainty analysis to convert unbounded outdoor environments into stable reconstruction regions and employs gradient-based scoring for precise camera--block allocation. This ensures structurally sound spatial partitioning, allowing each sub-block to focus on key observation regions while preventing degeneracy in global optimization.
    \item We contribute Pano360, the first large-scale panoramic dataset and benchmark for outdoor scene reconstruction, covering diverse commercial districts and urban natural landscapes. \ql{Extensive experiments on Pano360 demonstrate the effectiveness of PanoLOG, outperforming existing alternatives by a large margin.}
    
\end{itemize}

\section{Related Work}
\label{sec:related}

\subsection{Omnidirectional 3D Gaussian Splatting}

While 3D Gaussian Splatting (3DGS)~\citep{kerbl20233d} has established a new paradigm for real-time novel view synthesis, its core formulation assumes a perspective camera model, making it highly susceptible to the severe angular distortions inherent in equirectangular projections (ERP)~\citep{song2026unisharp}. To mitigate this, recent research has focused on mathematically adapting the 3DGS pipeline to omnidirectional image. Early efforts such as ODGS~\citep{lee2024odgs} address this by projecting Gaussians onto local tangent planes of the unit sphere. Subsequent methods, including OmniGS~\citep{li2025omnigs} and SC-OmniGS~\citep{huang2025sc}, derive closed-form derivatives to enable direct optimization in ERP space without intermediate rectification, with the latter further co-optimizing camera poses. More recently, 3DGEER~\citep{huang20263dgeer} generalized this approach by formulating an exact ray--Gaussian density integral applicable to arbitrary camera models.

Despite effectively resolving geometric distortions, existing panoramic 3DGS frameworks—along with parallel efforts in synthesis~\citep{chen2025splatter,zhang2025pansplat,lee2025omnisplat,wang2026cylindersplat} and generation~\citep{zhou2024dreamscene360,ma2024fastscene,huang2025scene4u}—primarily target localized manifolds. These methods face shared spatial constraints that hinder scalability in expansive outdoor environments, particularly when confronted with unbounded skies and low-texture regions where traditional spatial locality no longer holds. This gap underscores the necessity for a framework capable of bridging omnidirectional projection with large-scale scene management.

\subsection{Large-Scale Scene Reconstruction}

Scaling 3DGS to expansive environments introduces compounding hardware and optimization bottlenecks. To overcome GPU memory limits, the community has largely adopted divide-and-conquer strategies, progressively leveraging the inherent spatial constraints of pinhole cameras. For instance, VastGaussian~\citep{lin2024vastgaussian} partitions the scene into cells using an airspace-aware visibility criterion, while CityGaussian~\citep{liu2024citygaussian} utilizes camera frustums combined with rendering similarity metrics to assign cameras to specific spatial blocks. DOGS~\citep{chen2024dogs} further accelerates this paradigm via distributed ADMM-based optimization, achieving over 6$\times$ training speedup. Building upon these block-wise foundations, other works enhance scalability through hierarchical level-of-detail (LoD) structures~\citep{kerbl2024hierarchical,kulhanek2025lodge} or geometric constraints~\citep{liu2024citygaussianv2,chen2024gigags} for smoother large-scale navigation. Horizon-GS~\citep{jiang2025horizon} further unifies aerial and street-view reconstruction via a coarse-to-fine training strategy.

The omnipresent visibility of panoramic images renders traditional partitioning non-discriminative, causing block-wise pipelines to degenerate into global optimization and necessitating novel parallax- and gradient-guided strategies.

\section{Method}
\label{sec:method}

We present PanoLOG, a two-stage coarse-to-fine framework for large-scale panoramic 3DGS reconstruction.
We first introduce \ql{the base structure and loss design for each block (\cref{sec:prelim})}, then detail the two-stage training pipeline (\cref{sec:pipeline}): global coarse training (Stage~I) followed by block-wise refinement (Stage~II).

\subsection{Base Structure}
\label{sec:prelim}

Each spatial block shares a unified base structure: a panoramic 3DGS renderer with ERP projection, an explicit sky sphere, and panoramic depth supervision, all optimized under a common loss. We first describe the rendering formulation, then detail each auxiliary component.

\begin{figure}[t!]
\centering
\captionsetup{width=\textwidth}
\includegraphics[width=\textwidth]{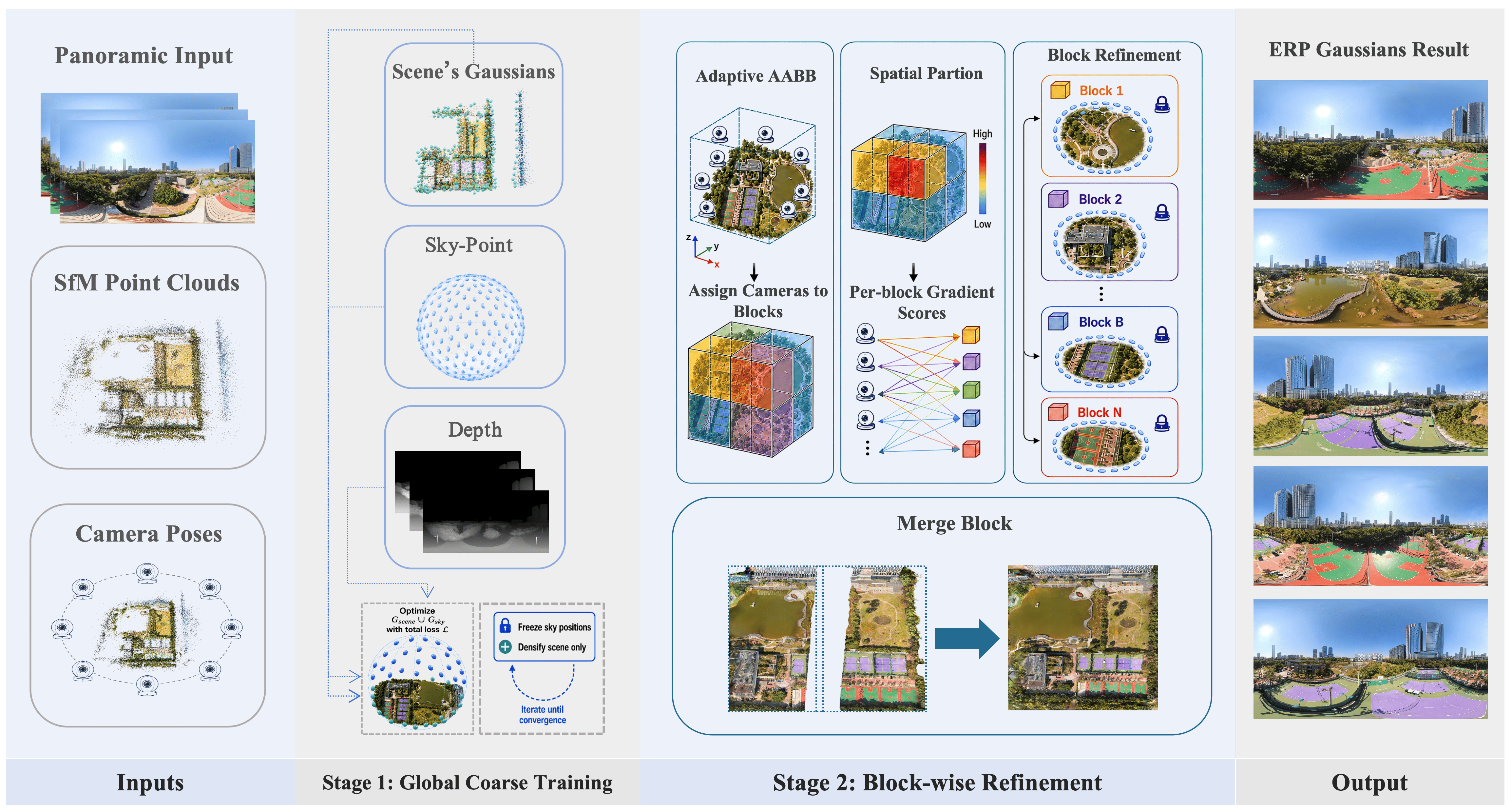}
\caption{Overview of the PanoLOG training pipeline. \textbf{Stage I} optimizes scene, sky, and depth priors to establish stable coarse geometry. \textbf{Stage II} introduces G$^2$PS for geometry-based spatial partitioning and gradient-driven camera--block allocation. Each block is then refined in parallel with frozen sky Gaussians and merged to produce the final high-quality reconstruction.}
\label{fig:pipeline}
\end{figure}

\paragraph{Panoramic 3DGS}

Inspired by~\citet{kerbl20233d}, we represent a scene with $N$ anisotropic 3D Gaussians $\mathcal{G}=\{g_i\}_{i=1}^N$, each parameterized by a mean $\boldsymbol{\mu}_i\!\in\!\mathbb{R}^3$, covariance $\mathbf{\Sigma}_i\!\in\!\mathbb{R}^{3\times 3}$, color $\mathbf{c}_i\!\in\!\mathbb{R}^3$, and opacity $o_i\!\in\!(0,1]$.
The rendered pixel color $\hat{C}(\mathbf{u})$ is computed via front-to-back alpha compositing:
\begin{equation}
\hat{C}(\mathbf{u}) = \sum_{i\in\mathcal{N}} \mathbf{c}_i\,\alpha_i \prod_{j=1}^{i-1}(1-\alpha_j), \quad \alpha_i = o_i \cdot G_{2\mathrm{D},i}(\mathbf{u}),
\label{eq:render}
\end{equation}
where $G_{2\mathrm{D},i}(\mathbf{u})$ is the projected 2D Gaussian at pixel $\mathbf{u}$.

To handle the full $360^{\circ}$ field of view, we adopt equirectangular projection (ERP).
For a 3D point $\mathbf{X}\!=\!(x,y,z)^\top$ in camera coordinates, its spherical coordinates $(\theta, \phi) = ( \mathrm{atan2}(x, z),\; \arcsin(y/\|\mathbf{X}\|))$ are mapped to pixel coordinates $\mathbf{u} = \boldsymbol{\pi}(\mathbf{X})$.
The 2D splatting covariance is obtained by projecting the 3D covariance through the ERP Jacobian $\mathbf{J} = \partial\boldsymbol{\pi}/\partial\mathbf{X}$:
\begin{equation}
\boldsymbol{\Sigma}_{2\mathrm{D}} = \mathbf{J}\,\mathbf{W}\,\boldsymbol{\Sigma}\,\mathbf{W}^\top\mathbf{J}^\top,
\label{eq:cov2d}
\end{equation}
where $\mathbf{W}$ is the world-to-camera rotation and $\boldsymbol{\Sigma} = \mathbf{R}\mathbf{S}\mathbf{S}^\top\mathbf{R}^\top$. Unlike the pinhole pipeline, $\boldsymbol{\Sigma}_{2\mathrm{D}}$ is computed via the nonlinear ERP Jacobian rather than a perspective projection matrix. The full derivation is provided in \cref{sec:app_erp}.

% \ql{\subsection{Base Structure}}
\paragraph{Auxiliary Design}
\label{sec:base}
Our base representation comprises \emph{explicit sky Gaussians} for modeling unbounded sky regions and \emph{standard 3D Gaussians} for near-field scene geometry, complemented by panoramic monocular depth priors and a unified optimization objective.
\yh{During block-wise refinement (Stage~II), each block inherits the sky Gaussians trained in Stage~I as a frozen global model, while the scene Gaussians are loaded as initialization and further optimized locally.}

\textit{Explicit Sky Sphere.}
Sky regions lack valid SfM geometry, causing near-field Gaussians to drift outward and produce floater artifacts. We initialize $N_{\mathrm{sky}}$ dedicated sky Gaussians on a distant sphere of radius $R_{\mathrm{sky}} = \kappa \cdot r_{\mathrm{scene}}$ ($\kappa\!=\!10$), where $r_{\mathrm{scene}} = \max_i \|\mathbf{x}_i - \bar{\mathbf{C}}\|$ is the maximum extent of the SfM point cloud from the camera centroid $\bar{\mathbf{C}}$. Positions are initialized via uniform spherical sampling:
\begin{equation}
\mathbf{p}_i^{\mathrm{sky}} = \bar{\mathbf{C}} + R_{\mathrm{sky}} \begin{pmatrix} \cos\theta_i\sin\phi_i \\ \sin\theta_i\sin\phi_i \\ \cos\phi_i \end{pmatrix}.
\label{eq:sky_pos}
\end{equation}

\textit{Panoramic Depth Supervision.}
\label{sec:depth}
ERP polar regions suffer from severe pixel stretching that degrades SfM triangulation, leaving the initial point cloud sparse and unreliable.
We employ DAP~\citep{lin2025depth}, a depth estimator natively supporting ERP input, to generate monocular inverse depth maps aligned to SfM sparse depth via per-image affine parameters:
\begin{equation}
\tilde{D}^{-1}_k = s_k \cdot D^{-1}_{\mathrm{mono},k} + o_k,
\label{eq:depth_align}
\end{equation}
where we operate in \emph{radial} depth ($\|\mathbf{X}\|$) rather than z-axis depth, as it is geometrically natural for omnidirectional cameras. The depth loss is:
\begin{equation}
\mathcal{L}_{\mathrm{depth}} = w(t) \cdot \frac{1}{|\Omega|}\sum_{\mathbf{u}\in\Omega}\left|\hat{D}^{-1}(\mathbf{u}) - \tilde{D}^{-1}_{\mathrm{mono}}(\mathbf{u})\right|,
\label{eq:depth_loss}
\end{equation}
where $\Omega$ is the set of valid pixels and $w(t)$ is an exponentially decaying weight that provides strong geometric guidance early while avoiding depth estimation errors from limiting final quality (schedule in \cref{sec:app_depth}).

\paragraph{Unified Optimization Objective.}
The overall training loss combines photometric and depth supervision:
\begin{equation}
\mathcal{L} = (1-\lambda_{\mathrm{ssim}})\,\mathcal{L}_1 + \lambda_{\mathrm{ssim}}\,\mathcal{L}_{\mathrm{D\text{-}SSIM}} + \mathcal{L}_{\mathrm{depth}},
\label{eq:total_loss}
\end{equation}
where $\mathcal{L}_1$ and $\mathcal{L}_{\mathrm{D\text{-}SSIM}}$ are the $\ell_1$ photometric loss and structural dissimilarity loss. This objective is shared across both training stages.

\subsection{Training Pipeline}
\label{sec:pipeline}

With the base representation defined above, we describe the two-stage procedure. Stage~I performs global coarse optimization to establish reliable geometry for partitioning; Stage~II refines each spatial block independently to recover fine-grained detail.

\subsubsection{Stage~I: Global Coarse Training}
\label{sec:coarse}

Directly partitioning raw SfM point clouds leads to floaters in out-of-block regions and unreliable camera--block allocation due to low-fidelity renderings.
We therefore first perform a global coarse optimization over all input panoramas to generate stable geometric priors.
Unlike prior frameworks~\citep{liu2024citygaussian} that rely solely on photometric supervision, we incorporate panoramic monocular depth priors (\cref{sec:depth}) from the outset, providing robust geometric anchoring critical for the subsequent gradient-based partitioning (\cref{sec:g2ps}).

During this stage, sky Gaussian position gradients are zeroed to prevent drift into the near field, while their appearance parameters (color, rotation, opacity) are optimized normally; sky Gaussians are also excluded from all densification operations.

\subsubsection{Stage~II: Block-wise Refinement}
\label{sec:refine}

\paragraph{Geometry and Gradient-based Partitioning Strategy (G$^2$PS).}
\label{sec:g2ps}
Equirectangular projection fundamentally violates the spatial visibility locality assumed by existing large-scale reconstruction frameworks: whereas conventional pinhole setups allow effective frustum-based partitioning because each block is observed by only a subset of cameras, the $360^{\circ}$ observation range of panoramic scenes couples all spatial regions across viewpoints, rendering such strategies indiscriminate and degenerating into global optimization.
We propose G$^2$PS, which addresses this through two complementary mechanisms: \emph{geometry-based spatial partitioning} that constructs a balanced block decomposition, and \emph{gradient-based camera--block allocation} that assigns viewpoints according to their actual observational contributions.

\textbf{\textit{Geometry-based Spatial Partitioning.}}
G$^2$PS constructs an adaptive axis-aligned bounding box (AABB) to define the effective reconstruction volume. Directly applying uniform grid partitioning in unbounded 3D space produces many nearly empty blocks with severe workload imbalance; we therefore contract the scene into a bounded AABB whose extent is determined by camera distribution and triangulation reliability.

We first compute the camera centroid $\bar{\mathbf{C}} = \frac{1}{K}\sum_{k}\mathbf{C}_k$ and the per-axis base radius $r_d = \max_{k}|\mathbf{C}_k^{(d)} - \bar{\mathbf{C}}^{(d)}|$ for $d\in\{x,y,z\}$, which encloses the camera trajectory. Since panoramic scene content extends well beyond the cameras, we expand the boundary according to the effective range of spherical feature triangulation. For ERP panoramas, matched features are lifted to unit bearing rays, and triangulation reliability is governed by the angular parallax $\gamma$. Under the small-parallax approximation, depth satisfies $z \sim b/\gamma$; by first-order error propagation, the relative depth uncertainty is $\sigma_z/z \approx (z/b)\,\sigma_\gamma$. Bounding this by a tolerance $\eta$ and defining the triangulation range factor $\rho_{\mathrm{tri}}=\eta/\sigma_{\gamma}$, we obtain $z_{\max} \approx b\,\rho_{\mathrm{tri}}$.

The expansion margin is defined as:
\begin{equation}
\mathrm{margin} = \hat{b}\cdot \rho_{\mathrm{tri}},
\label{eq:margin}
\end{equation}
with the representative baseline $\hat{b}$ robustly estimated as the median nearest-neighbor camera distance:
\begin{equation}
\hat{b} = \mathrm{median}_{k}\left(\min_{j\neq k}\|\mathbf{C}_k - \mathbf{C}_j\|\right).
\label{eq:baseline}
\end{equation}
The final AABB bounds are:
\begin{equation}
\mathbf{a}_{\min}^{(d)} = \bar{\mathbf{C}}^{(d)} - r_d - \mathrm{margin}, \quad \mathbf{a}_{\max}^{(d)} = \bar{\mathbf{C}}^{(d)} + r_d + \mathrm{margin}.
\label{eq:aabb}
\end{equation}
Gaussians extending beyond the AABB are contracted into the bounded region as prescribed by \citet{liu2024citygaussian}. The contracted cubic space is then uniformly partitioned into spatial blocks, yielding a balanced distribution of Gaussians.

\textbf{\textit{Gradient-based Camera-Block Allocation.}}
In the refinement stage, each block must receive sufficient training supervision. We observe that the rendering loss gradient with respect to Gaussian positions naturally reflects their importance to the current viewpoint: regions with large contributions exhibit large gradients, while distant or occluded regions have near-zero gradients. After Stage~I converges, we perform a single forward-backward pass over all training views to collect these gradients. For camera $k$, we compute the mean gradient magnitude over all Gaussians $\mathcal{G}_b$ within block $b$:
\begin{equation}
s_{k,b} = \frac{1}{|\mathcal{G}_b|}\sum_{g\in\mathcal{G}_b}\left\|\frac{\partial\mathcal{L}_k}{\partial\mathbf{x}_g}\right\|_1,
\label{eq:grad_score}
\end{equation}
where $\mathcal{L}_k$ denotes the rendering loss for camera $k$ and $\mathbf{x}_g$ is the position of Gaussian $g$. The camera--block assignment combines geometric membership with gradient-based importance:
\begin{equation}
\mathrm{assign}(k, b) = \underbrace{\mathbb{1}\!\left[\mathbf{C}_k \in \mathrm{Block}_b\right]}_{\text{geometric}} \;\lor\; \underbrace{\mathbb{1}\!\left[\frac{s_{k,b}}{\max_{b'}\, s_{k,b'}} > \tau_{\mathrm{grad}}\right]}_{\text{gradient-based}},
\label{eq:assign}
\end{equation}
where $\tau_{\mathrm{grad}}=0.8$ is the gradient ratio threshold. A camera is assigned to a block if it is either geometrically located within that block or its normalized gradient score exceeds the threshold, ensuring that viewpoints with significant observational contributions are included even when physically located in adjacent blocks.

\paragraph{Block-wise Optimization.}
\label{sec:block_opt}
Each block is initialized with the \emph{full} set of Gaussians from Stage~I and optimized independently using its assigned cameras. Sky Gaussians are fully frozen during this stage to maintain cross-block appearance consistency.

To adaptively determine each block's effective spatial extent, we leverage periodic opacity reset. After each reset, all scene Gaussian opacities are clamped to a low value $o_{\mathrm{reset}}$; only Gaussians receiving sufficient photometric gradients from the block's assigned cameras recover high opacity, while the rest remain transparent and are pruned. The retained Gaussian set for block $b$ is:
\begin{equation}
\tilde{\mathcal{G}}_b = \bigl\{\, g \in \mathcal{G}_{b} \;\big|\; \mathbf{x}^{\mathrm{ctr}}_g \in [\mathbf{l}_b,\mathbf{u}_b) \;\land\; o_g^{(b)} > o_{\min} \,\bigr\},
\label{eq:block_prune}
\end{equation}
where $\mathcal{G}_{b}$ is the Gaussian set after block $b$'s optimization, $\mathbf{x}^{\mathrm{ctr}}_g$ is the center of Gaussian $g$ in the contracted coordinate space, $[\mathbf{l}_b,\mathbf{u}_b)$ is the half-open axis-aligned cell of block $b$, $o_g^{(b)}$ is the converged opacity, and $o_{\min}$ is the pruning threshold. The half-open interval convention ensures each retained Gaussian is assigned to exactly one block, avoiding duplicate primitives in the final model. The merged model is:
\begin{equation} \mathcal{G}_{\mathrm{final}} = \mathcal{G}_{\mathrm{sky}} \;\cup\; \bigcup_{b=1}^{B} \tilde{\mathcal{G}}_b. \label{eq:merge} \end{equation}
Because adjacent blocks share overlapping camera sets (via \cref{eq:assign}), Gaussians near block boundaries receive consistent cross-boundary supervision, producing coherent geometry and appearance at block interfaces without explicit stitching.

\section{Experiments}
\label{sec:experiments}

\begin{figure*}[t!]
\centering
\includegraphics[width=\textwidth]{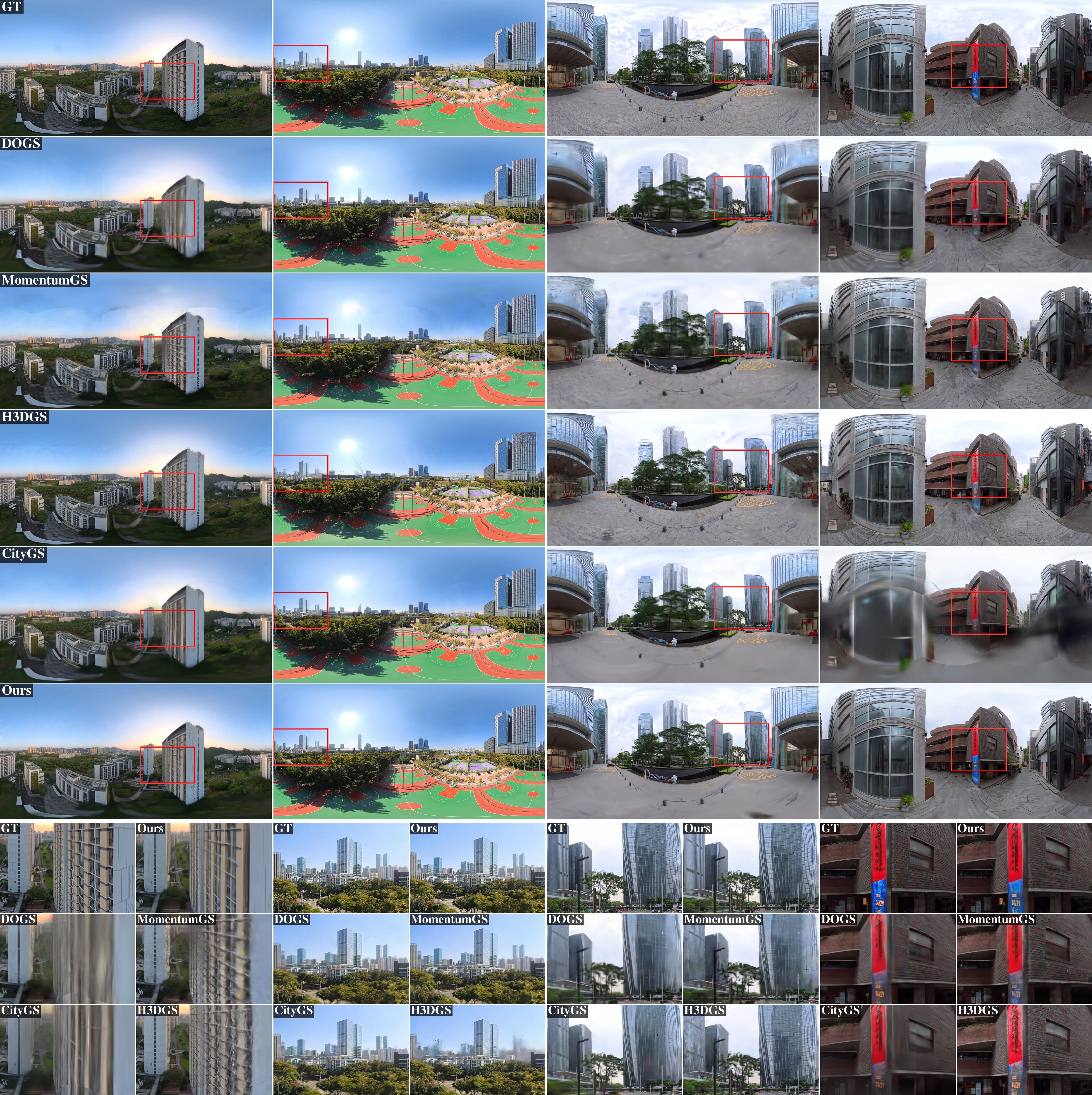}
\caption{We provide a comprehensive comparison between PanoLOG and four existing methods across diverse large-scale scenes. Results from both panoramic inputs and local details show that PanoLOG consistently outperforms the baselines. Notably, our method excels in reconstructing both distant regions and complex glass facades, which are typically challenging in urban environments.}
\label{fig:comparison_all}
\end{figure*}

\subsection{Experimental Setup}
\paragraph{Datasets.}
We introduce \textbf{Pano360}, the first large-scale outdoor panoramic reconstruction dataset, comprising 5,637 high-resolution (3840$\times$1920) images captured across four diverse urban environments. Pano360 consists of four subsets—NSC, NSK, BAX, and NSN—where NSC and NSK are collected by the A1 drone, while BAX and NSN are acquired with the X5 handheld camera. The dataset provides calibrated poses and sparse point clouds with a 7:1 train/test split. To demonstrate generalizability, we further evaluate on the public Ricoh360~\citep{choi2023balanced} and 360Roam~\citep{huang2022360roam} benchmarks. Additional dataset details (coverage, cropping protocol, and anonymization) are provided in \cref{supp:dataset}.

\paragraph{Baselines and Metrics.}
We compare against large-scale 3DGS methods---H3DGS~\citep{kerbl2024hierarchical}, CityGaussian~\citep{liu2024citygaussian}, DOGS~\citep{chen2024dogs}, and Momentum-GS~\citep{fan2025momentum}---by converting ERP images into six cubemap views with pinhole camera models. On public datasets, we additionally compare with panoramic 3DGS methods: OmniGS~\citep{li2025omnigs}, ODGS~\citep{lee2024odgs}, SpaGS~\citep{li2025spags}, and a cubemap-input 3DGS baseline. Metrics include PSNR, SSIM~\citep{wang2004image}, LPIPS-VGG~\citep{zhang2018unreasonable}, and model size.

\paragraph{Implementation Details.}
All methods use official implementations with recommended hyperparameters, are trained for 30,000 iterations, and use two partitions by default. All experiments run on a single NVIDIA RTX 4090 GPU (24 GB). Due to GPU memory limitations, H3DGS uses eight partitions on the BAX and NSN datasets. Full training configurations are provided in \cref{supp:impl_details}.

\subsection{Benchmarking Results}

\paragraph{Results on Pano360.}
\cref{tab:a1_results} and \cref{tab:x5_results} report quantitative results on the A1 drone and X5 handheld sub-scenes, respectively. PanoLOG achieves the best PSNR on three of four sub-scenes (NSC, BAX, NSN) and the best SSIM and LPIPS on NSK. On the X5 scenes, PanoLOG outperforms the strongest baseline by $0.64$\,dB on BAX and $1.16$\,dB on NSN, while our models are $2.9$--$7.5\times$ smaller than the strongest baseline. \cref{fig:comparison_all} provides qualitative comparisons; additional visual results are in \cref{supp:public_visual_root}.

\begin{table}[t]
\caption{Quantitative comparison on the A1 drone sub-scenes (NSC, NSK) of Pano360. Size is in MB. Our method achieves the best results while maintaining the smallest model size. {\color{red}\textbf{Bold red}} and {\color{orange}\textbf{bold orange}} denote the best performance and the largest model, respectively.}
\label{tab:a1_results}
\centering
\small
\resizebox{\textwidth}{!}{
\begin{tabular}{l cccc cccc}
\toprule
& \multicolumn{4}{c}{NSC} & \multicolumn{4}{c}{NSK} \\
\cmidrule(lr){2-5} \cmidrule(lr){6-9}
Method & PSNR$\uparrow$ & SSIM$\uparrow$ & LPIPS$\downarrow$ & Size$\downarrow$ & PSNR$\uparrow$ & SSIM$\uparrow$ & LPIPS$\downarrow$ & Size$\downarrow$ \\
\midrule
H3DGS~\citep{kerbl2024hierarchical} & 27.7787 & 0.8564 & 0.2457 & 1002.1 & 24.1480 & 0.8154 & 0.1934 & \textbf{\textcolor{orange}{1843.2}} \\
CityGaussian~\citep{liu2024citygaussian} & 27.7340 & 0.8453 & 0.2609 & 523.7 & \textbf{\textcolor{red}{24.8270}} & 0.8176 & 0.1983 & 1126.4 \\
DOGS~\citep{chen2024dogs} & 26.8486 & 0.8186 & 0.2769 & \textbf{\textcolor{orange}{1024.0}} & 24.4606 & 0.7980 & 0.2174 & 1536.0 \\
Momentum-GS~\citep{fan2025momentum} & 26.4568 & 0.8311 & 0.2625 & 802.5 & 23.9123 & 0.7979 & 0.1986 & 1024.0 \\
\textbf{Ours} & \textbf{\textcolor{red}{28.1838}} & \textbf{\textcolor{red}{0.8594}} & \textbf{\textcolor{red}{0.2435}} & 463.5 & 24.6387 & \textbf{\textcolor{red}{0.8243}} & \textbf{\textcolor{red}{0.1916}} & 544.0 \\
\bottomrule
\end{tabular}
}
\end{table}

\begin{table}[t]
\vspace{-1mm}
\caption{Quantitative comparison on the X5 handheld sub-scenes (BAX, NSN) of Pano360. Size is in MB. Although our model size increases in these scenes, it remains below the largest baseline; compared with methods of similar size, our results consistently surpass them, and we achieve the best performance on several metrics.}
\label{tab:x5_results}
\centering
\small
\resizebox{\textwidth}{!}{
\begin{tabular}{l cccc cccc}
\toprule
& \multicolumn{4}{c}{BAX} & \multicolumn{4}{c}{NSN} \\
\cmidrule(lr){2-5} \cmidrule(lr){6-9}
Method & PSNR$\uparrow$ & SSIM$\uparrow$ & LPIPS$\downarrow$ & Size$\downarrow$ & PSNR$\uparrow$ & SSIM$\uparrow$ & LPIPS$\downarrow$ & Size$\downarrow$ \\
\midrule
H3DGS~\citep{kerbl2024hierarchical} & 20.7104 & 0.6626 & \textbf{\textcolor{red}{0.3936}} & \textbf{\textcolor{orange}{7577.6}} & 23.4464 & \textbf{\textcolor{red}{0.7530}} & \textbf{\textcolor{red}{0.3018}} & \textbf{\textcolor{orange}{5734.4}} \\
CityGaussian~\citep{liu2024citygaussian} & 19.8361 & 0.5995 & 0.4902 & 418.5 & 21.9492 & 0.6729 & 0.4221 & 526.6 \\
DOGS~\citep{chen2024dogs} & 19.6094 & 0.5745 & 0.5203 & 671.8 & 22.6743 & 0.6587 & 0.4454 & 774.9 \\
Momentum-GS~\citep{fan2025momentum} & 20.0993 & 0.5999 & 0.4896 & 1024.0 & 22.9746 & 0.7052 & 0.3703 & 1331.2 \\
\textbf{Ours} & \textbf{\textcolor{red}{21.3468}} & \textbf{\textcolor{red}{0.6656}} & 0.4169 & 1331.2 & \textbf{\textcolor{red}{24.6095}} & 0.7508 & 0.3347 & 766.8 \\
\bottomrule
\end{tabular}
}
\end{table}

\paragraph{Results on Public Panoramic Datasets.}
As shown in \cref{tab:public_results}, PanoLOG achieves the best PSNR, SSIM, and LPIPS on both Ricoh360 and 360Roam, outperforming all panoramic baselines (OmniGS, ODGS, SpaGS) and the cubemap 3DGS baseline across all metrics. This demonstrates that the panoramic-native design of PanoLOG generalizes beyond our own benchmark. Per-scene breakdowns and qualitative comparisons are provided in \cref{supp:public_visual}.

\begin{table}[t!]
% \vspace{-4mm}
\caption{PanoLOG outperforms four state-of-the-art methods across two public benchmarks, achieving superior results in all metrics.}
\label{tab:public_results}
\centering
\begin{tabular}{l ccc ccc}
\toprule
& \multicolumn{3}{c}{Ricoh360} & \multicolumn{3}{c}{360Roam} \\
\cmidrule(lr){2-4} \cmidrule(lr){5-7}
Method & PSNR$\uparrow$ & SSIM$\uparrow$ & LPIPS$\downarrow$ & PSNR$\uparrow$ & SSIM$\uparrow$ & LPIPS$\downarrow$ \\
\midrule
OmniGS~\citep{li2025omnigs} & 26.00 & 0.828 & 0.210 & 24.93 & 0.800 & 0.253 \\
3DGS~\citep{kerbl20233d} & 26.26 & 0.825 & 0.225 & 24.98 & 0.800 & 0.271 \\
ODGS~\citep{lee2024odgs} & 22.71 & 0.748 & 0.326 & 19.74 & 0.627 & 0.476 \\
SpaGS~\citep{li2025spags} & 26.11  & 0.832  & 0.243 & 25.45  & 0.814 & 0.223 \\
\textbf{Ours} & \textbf{\textcolor{red}{26.48}} & \textbf{\textcolor{red}{0.845}} & \textbf{\textcolor{red}{0.183}} & \textbf{\textcolor{red}{25.83}} & \textbf{\textcolor{red}{0.821}} & \textbf{\textcolor{red}{0.222}} \\
\bottomrule
\end{tabular}
\end{table}

\begin{figure*}[t]
\centering
\includegraphics[width=\textwidth]{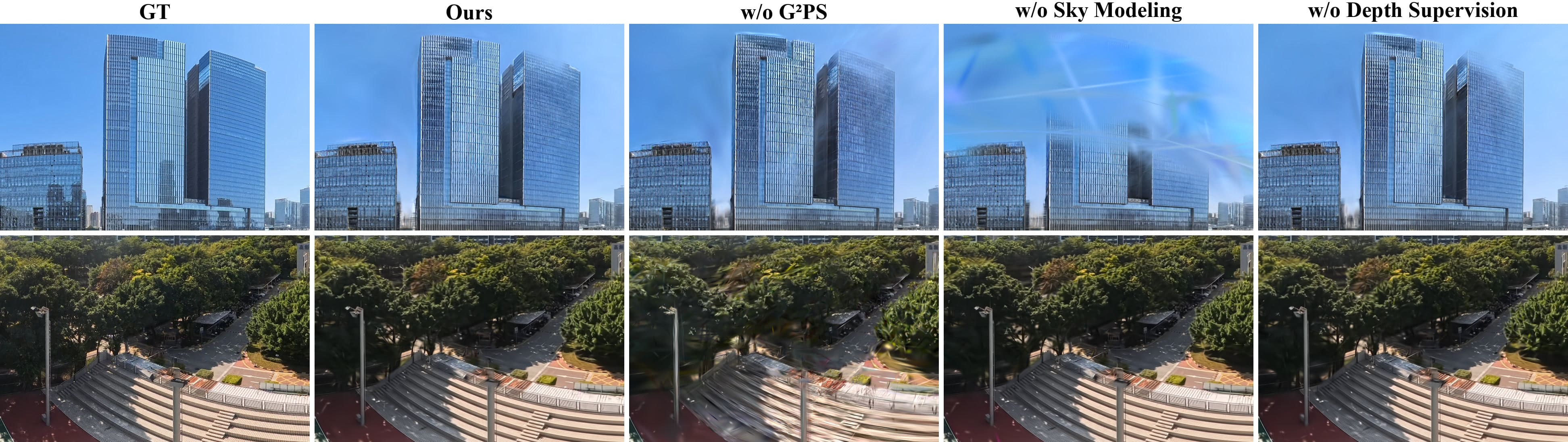}
\caption{Ablation study on two representative scenes(from NSK dataset). We compare our full model against the ground truth (GT) and variants excluding G$^2$PS, explicit sky optimization, and depth supervision. The integration of these components significantly enhances rendering quality for infinite sky, glass facades, and dense vegetation.}
\label{fig:ablation_fig}
\end{figure*}

\subsection{Ablation Study}

We ablate three key components: (1)~replacing G$^2$PS with position-only partitioning, (2)~removing the explicit sky sphere, and (3)~removing the depth supervision. Results are in \cref{tab:ablation} and \cref{fig:ablation_fig}. Replacing G$^2$PS reduces PSNR by $0.51$\,dB, confirming gradient-based allocation is essential. Without the sky sphere, floaters proliferate due to unconstrained Gaussian drift. Removing depth weakens geometric supervision in polar and distant areas.

\begin{table}[t!]
\begin{minipage}[t]{0.50\linewidth}
\caption{Ablation study on key components of PanoLOG. Sup. denotes Supervision.}
\label{tab:ablation}
\centering
\small
\setlength{\tabcolsep}{3pt}
\begin{tabular}{@{}lcccc@{}}
\toprule
Setting & PSNR$\uparrow$ & SSIM$\uparrow$ & LPIPS$\downarrow$ & Size$\downarrow$ \\
\midrule
w/o G$^2$PS & 24.14 & 0.816 & 0.194 & \textbf{580M} \\
w/o Sky Sphere & 24.15 & 0.822 & 0.195 & 683M \\
w/o Depth Sup. & 24.61 & 0.824 & 0.191 & 588M \\
Full Model & \textbf{24.65} & \textbf{0.827} & \textbf{0.186} & 667M \\
\bottomrule
\end{tabular}
\end{minipage}
\hfill
\begin{minipage}[t]{0.47\linewidth}
\caption{Effect of $\tau_{\mathrm{grad}}$ on quality and model size. $\tau_{\mathrm{grad}}=0.8$ best balances quality and compactness.}
\label{tab:threshold}
\centering
\small
\setlength{\tabcolsep}{3pt}
\begin{tabular}{@{}lcccc@{}}
\toprule
Setting & PSNR$\uparrow$ & SSIM$\uparrow$ & LPIPS$\downarrow$ & Size$\downarrow$ \\
\midrule
$\tau_{\mathrm{grad}} = 0.7$ & 24.61 & 0.822 & 0.192 & 699M \\
$\tau_{\mathrm{grad}} = 0.8$ & \textbf{24.65} & \textbf{0.827} & \textbf{0.186} & \textbf{667M} \\
$\tau_{\mathrm{grad}} = 0.9$ & 24.42 & 0.823 & 0.190 & 654M \\
\bottomrule
\end{tabular}
\end{minipage}
\end{table}

\subsection{Analysis of Partition Threshold}

We vary $\tau_{\mathrm{grad}} \in \{0.7, 0.8, 0.9\}$ in the camera--block allocation (\cref{eq:assign}). As shown in \cref{tab:threshold}, the default $\tau_{\mathrm{grad}}=0.8$ achieves the best trade-off. Further analysis is provided in \cref{supp:threshold}.

\section{Conclusion}
We presented PanoLOG, a two-stage coarse-to-fine framework for large-scale outdoor panoramic 3DGS reconstruction.
The central contribution is G$^2$PS, which resolves the omnipresent visibility problem inherent to panoramic scenes through geometry-based spatial partitioning and gradient-based camera--block allocation.
Complementary components---explicit sky sphere modeling and panoramic monocular depth supervision---further stabilize reconstruction in unbounded and geometrically challenging regions.
We also introduced Pano360, the first large-scale panoramic outdoor benchmark, comprising 5,637 high-resolution images across four diverse urban environments.
Extensive experiments on Pano360 and public benchmarks demonstrate that PanoLOG achieves state-of-the-art rendering quality while producing models $2.2$--$7.5\times$ smaller than H3DGS.
These results confirm that panoramic-aware partitioning is essential for scalable, high-fidelity scene reconstruction from omnidirectional image.

\bibliographystyle{plainnat}
\bibliography{example_paper}

%%%%%%%%%%%%%%%%%%%%%%%%%%%%%%%%%%%%%%%%%%%%%%%%%%%%%%%%%%%%
\newpage
% --- 建议在主文件或附录开头添加的宏包 (若尚未添加) ---
% \usepackage{booktabs} % 用于高质量表格
% \usepackage{amsmath, amssymb} % 数学公式
% \usepackage{caption} % 图表标题控制

\appendix
\section*{Appendix}
\addcontentsline{toc}{section}{Appendix}

% --- 计数器重置：使附录中的图表编号变为 A.1, A.2 等 ---
\renewcommand{\thesection}{\Alph{section}}
\renewcommand{\thetable}{\Alph{section}.\arabic{table}}
\renewcommand{\thefigure}{\Alph{section}.\arabic{figure}}
\setcounter{table}{0}
\setcounter{figure}{0}

\section{Additional Implementation Details}
\label{supp:impl_details}

\begin{itemize}
    \item \textbf{Training Configuration:} Both Stage~I (global coarse) and Stage~II (block-wise refinement) are trained for 30{,}000 iterations following the standard 3DGS densification schedule. All methods use identical learning rates: $1.6\times10^{-4}$ for Gaussian positions and $2.5\times10^{-3}$ for spherical harmonics coefficients. The densification interval is 100 iterations with a gradient threshold of $2\times10^{-4}$. Opacity reset is performed every 3{,}000 iterations during block-wise refinement.
    \item \textbf{Partitioning Parameters:} The triangulation range factor is set to $\rho_{\mathrm{tri}}=5$ across all scenes. The gradient ratio threshold is $\tau_{\mathrm{grad}}=0.8$. The sky sphere multiplier is $\kappa=10$ with $N_{\mathrm{sky}}=100{,}000$ sky Gaussians.
    \item \textbf{Depth Supervision:} We use DAP~\citep{lin2025depth} to generate monocular inverse depth maps. The depth weight schedule follows \cref{eq:depth_weight} with $w_0=0.5$ and $w_T=0.01$.
    \item \textbf{Hardware and Software:} All experiments run on a single NVIDIA RTX 4090 GPU (24\,GB VRAM). The codebase is built on PyTorch 2.1 with custom CUDA kernels for ERP-based Gaussian rasterization.
\end{itemize}

\section{ERP Projection and Gaussian Rendering Details}
\label{sec:app_erp}

This section provides the full derivation of the equirectangular projection pipeline summarized in \cref{sec:prelim}.

\paragraph{ERP Pixel Mapping.}
For a 3D point $\mathbf{X}=(x,y,z)^\top$ in camera coordinates with spherical coordinates $(\theta, \phi)$, the ERP pixel coordinates $\mathbf{u}=(u,v)^\top$ are:
\begin{equation}
\mathbf{u} = \boldsymbol{\pi}(\mathbf{X}) = \begin{pmatrix} \frac{W}{2\pi}\,\mathrm{atan2}(x,z) + \frac{W}{2} \\[4pt] \frac{H}{\pi}\,\arcsin\!\left(\frac{y}{\|\mathbf{X}\|}\right) + \frac{H}{2} \end{pmatrix},
\label{eq:erp_proj}
\end{equation}
where $W$ and $H$ denote the panoramic image width and height.

\paragraph{ERP Jacobian.}
The Jacobian $\mathbf{J}\in\mathbb{R}^{2\times 3}$ of the projection $\boldsymbol{\pi}$ at point $\mathbf{X}$ is:
\begin{equation}
\mathbf{J} = \frac{\partial \boldsymbol{\pi}}{\partial \mathbf{X}} = \begin{pmatrix} \frac{W}{2\pi}\cdot\frac{z}{x^2+z^2} & 0 & -\frac{W}{2\pi}\cdot\frac{x}{x^2+z^2} \\[6pt] -\frac{H}{\pi}\cdot\frac{xy}{\|\mathbf{X}\|^2\sqrt{x^2+z^2}} & \frac{H}{\pi}\cdot\frac{\sqrt{x^2+z^2}}{\|\mathbf{X}\|^2} & -\frac{H}{\pi}\cdot\frac{zy}{\|\mathbf{X}\|^2\sqrt{x^2+z^2}} \end{pmatrix}.
\label{eq:jacobian}
\end{equation}

\paragraph{2D Gaussian Evaluation.}
Given the projected center $\mathbf{u}_s$ obtained via \cref{eq:erp_proj} and a sampling pixel $\mathbf{u}$, the 2D Gaussian is evaluated as:
\begin{equation}
G_{2\mathrm{D}}(\mathbf{u}) = \exp\!\left\{ -\frac{1}{2}(\mathbf{u}-\mathbf{u}_s)^\top \boldsymbol{\Sigma}_{2\mathrm{D}}^{-1}\, (\mathbf{u}-\mathbf{u}_s) \right\},
\label{eq:2d_gauss}
\end{equation}
where $\boldsymbol{\Sigma}_{2\mathrm{D}}$ is the projected 2D covariance defined in \cref{eq:cov2d}. The panoramic rendering then follows \cref{eq:render} with this ERP-specific $G_{2\mathrm{D}}$.

\section{Depth Weight Decay Schedule}
\label{sec:app_depth}

The time-decaying weight $w(t)$ in the depth loss (\cref{eq:depth_loss}) follows an exponential schedule:
\begin{equation}
w(t) = w_0 \cdot \left(\frac{w_T}{w_0}\right)^{t/T},
\label{eq:depth_weight}
\end{equation}
where $w_0$ and $w_T$ are the initial and final depth weights, and $T$ is the total number of training iterations. This yields $w(0)=w_0$ and $w(T)=w_T$ exactly. The exponential form ensures rapid early guidance, steering Gaussians from sparse SfM initialization toward geometrically plausible positions, while smoothly transitioning to photometric-dominated optimization without abrupt weight changes.

\section{Block-wise Optimization Details}
\label{sec:app_block}

This section supplements \cref{sec:g2ps,sec:block_opt} with additional design rationale.

\paragraph{Full Initialization.}
Each block is initialized by loading the \emph{full} set of Gaussians from Stage~I rather than only those geometrically inside the block. This prevents newly densified Gaussians near block boundaries from replacing pruned structures that were well-optimized during coarse training, which would otherwise introduce artifacts.

\paragraph{Soft-Boundary via Opacity Reset.}
To adaptively determine each block's effective spatial extent, we leverage periodic opacity reset. The reset operation periodically clamps all scene Gaussian opacities to a low value $o_{\mathrm{reset}}$, after which only those receiving sufficient photometric gradients from the block's assigned cameras recover high opacity through gradient-driven updates. Insufficiently observed Gaussians remain transparent and are subsequently pruned. The half-open interval $[\mathbf{l}_b,\mathbf{u}_b)$ ensures each retained Gaussian is uniquely assigned to one block, avoiding duplicate primitives in the final merged model. Because adjacent blocks share overlapping camera sets (via \cref{eq:assign}), Gaussians near block boundaries receive consistent cross-boundary supervision, yielding coherent geometry and appearance at block interfaces without explicit stitching.

\paragraph{Sky Gaussian Freezing.}
All sky Gaussian parameter gradients are zeroed during block-wise refinement. Since each block observes only a partial set of viewpoints, allowing sky Gaussians to update independently across blocks would cause inconsistent appearance drift. Full freezing maintains cross-block sky consistency.

\paragraph{Depth Supervision in Block Refinement.}
The depth supervision introduced in \cref{sec:depth} remains active during block refinement, where its role becomes even more critical: each block utilizes only a subset of camera viewpoints, further weakening the geometric constraints available from SfM. The monocular depth prior compensates for this reduced multi-view coverage, stabilizing optimization in geometrically under-constrained regions.

\section{Dataset Details}
\label{supp:dataset}

\textbf{Pano360} spans over 2 million $\text{m}^2$ across four diverse urban environments. All subset names (NSC, NSK, BAX, NSN) are anonymized identifiers to avoid disclosing specific geographic locations. Throughout, A1 refers to the Antigravity A1 panoramic UAV, and X5 denotes the Insta360 X5 panoramic camera.

\paragraph{Capture Protocol.}
NSC and NSK are captured by the A1 drone at an elevated viewpoint, which inherently avoids ground-level pedestrians. BAX and NSN are acquired with the X5 handheld camera at ground level, where the camera operator and other pedestrians are persistently visible near the image nadir.

\paragraph{Evaluation Cropping.}
Since the X5 camera is handheld, dynamic figures appear near the bottom of the ERP image. To ensure fair comparison, when reporting quantitative metrics on the X5 sub-scenes, we uniformly crop approximately 10\% of the ERP image height from the bottom for all evaluated methods (including baselines), thereby excluding dynamic figures from the evaluation region. No such cropping is applied to the A1 sequences.

\section{Analysis of Partition Threshold}
\label{supp:threshold}

We vary the gradient ratio threshold $\tau_{\mathrm{grad}} \in \{0.7, 0.8, 0.9\}$ in the camera--block allocation (\cref{eq:assign}). Lower thresholds assign more cameras per block: at $\tau_{\mathrm{grad}}=0.7$, each block receives broader supervision, which slightly improves rendering quality but increases model size due to more overlapping Gaussians across blocks. Higher thresholds produce more compact models by restricting camera assignments, but risk leaving boundary regions under-supervised. At $\tau_{\mathrm{grad}}=0.9$, the restriction becomes excessive, reducing quality on certain blocks. The default $\tau_{\mathrm{grad}}=0.8$ achieves the best balance between rendering quality and model compactness across all tested scenes.

\section{Additional Experimental Results}
\label{supp:extra_results}

\subsection{Additional Qualitative Results on Pano360}
\label{supp:public_visual_root}
\begin{figure*}[t]
    \centering
    \includegraphics[width=\textwidth]
    {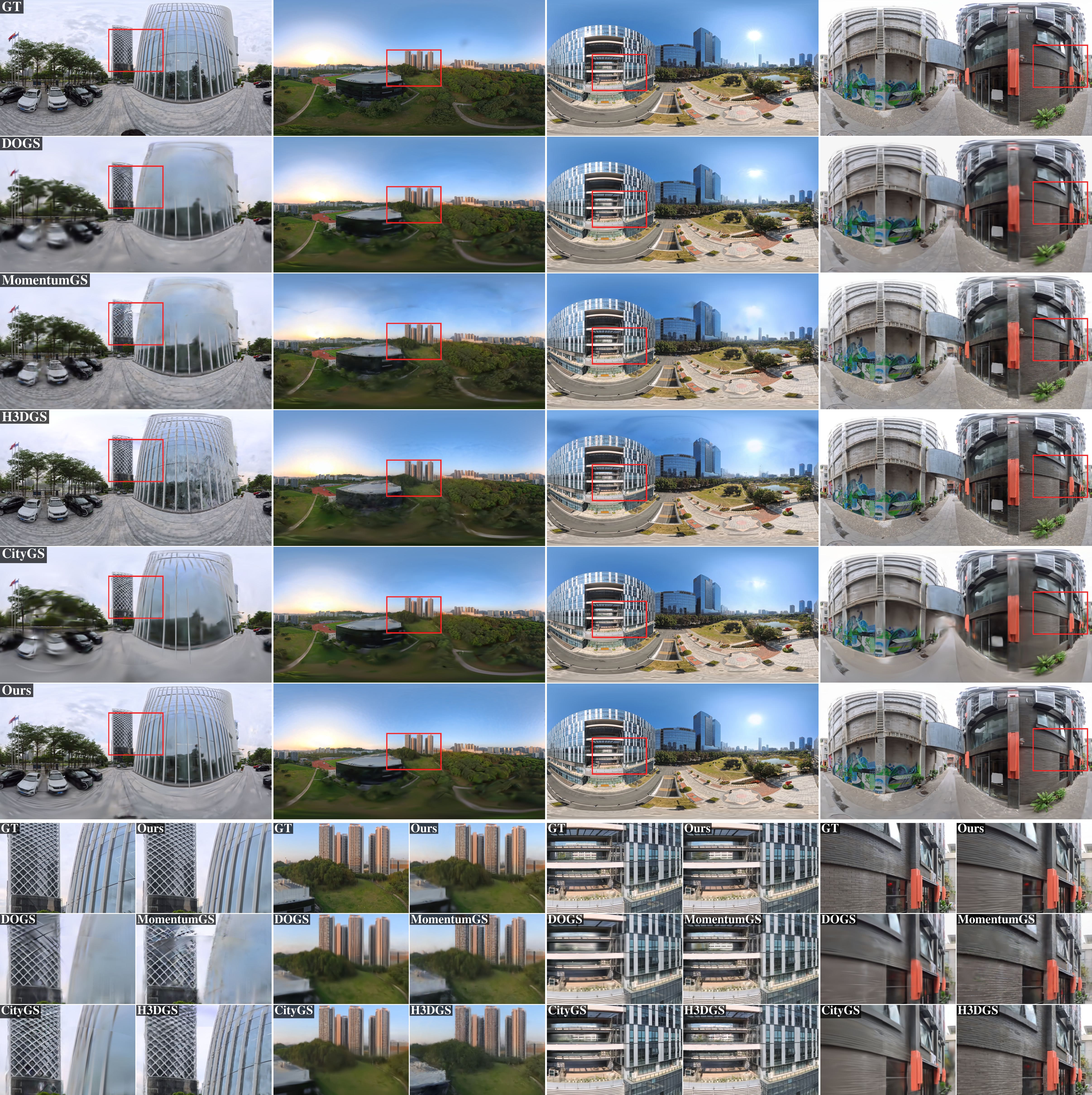}
    \caption{Additional qualitative comparisons on Pano360. We show more rendering results across diverse scenes to complement \cref{fig:comparison_all} in the main text.}
    \label{fig:supp_pano360_visual}
\end{figure*}

\subsection{Qualitative Results on Public Panoramic Datasets}
\label{supp:public_visual}

\begin{figure*}[h]
    \centering
    \includegraphics[width=\textwidth]{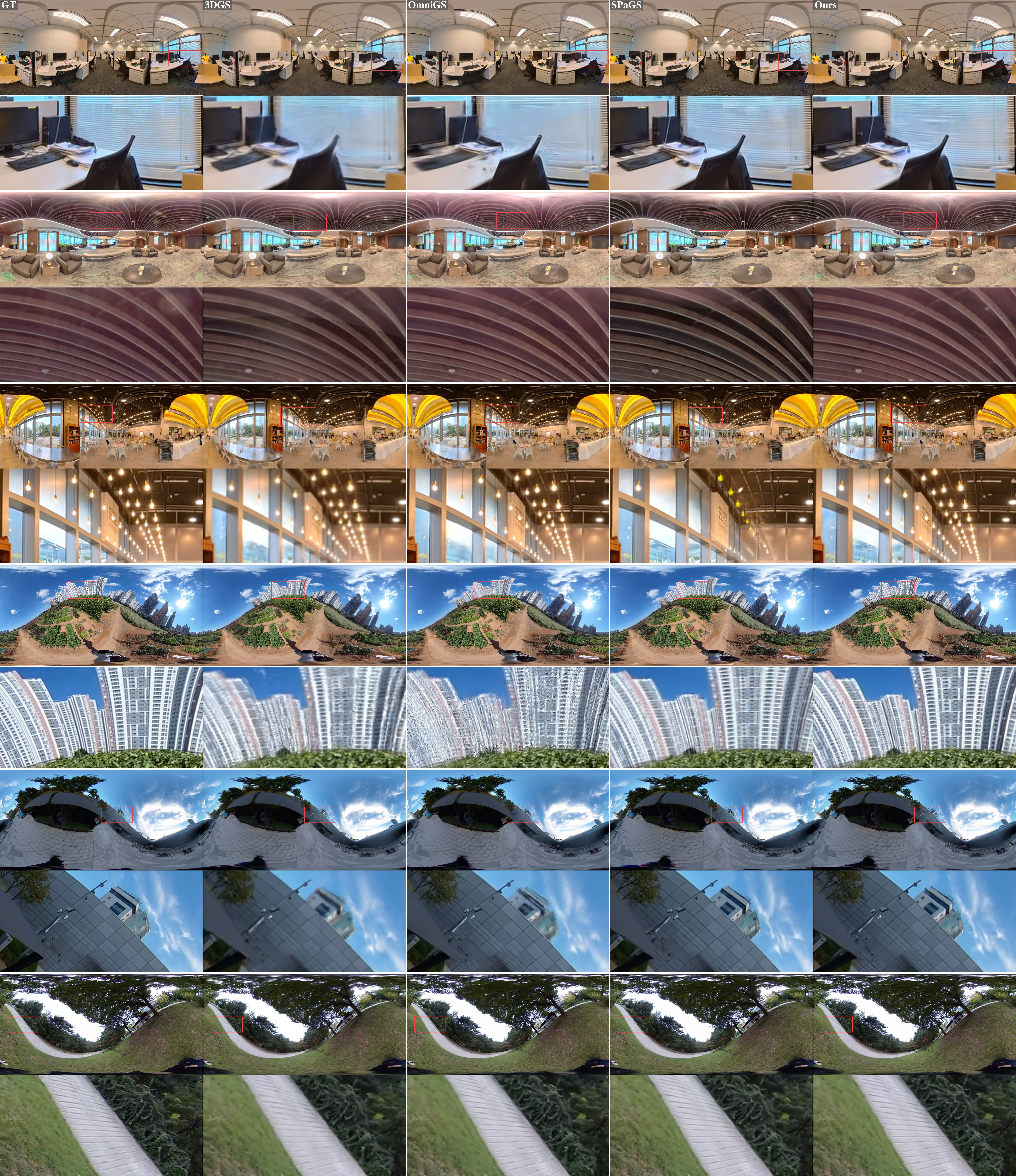}
    \caption{Additional qualitative results on 360Roam and Ricoh360. Our method consistently produces sharper textures and fewer artifacts across diverse indoor and outdoor scenes.}
    \label{fig:supp_public_visual}
\end{figure*}

\section{Broader Impact and Limitations}
\subsection{Broader Impact}
This work introduces \textit{PanoLOG}, a scalable framework for large-scale outdoor panoramic reconstruction. By enhancing the efficiency and fidelity of 3D environment modeling, it has the potential to advance fields such as autonomous driving simulation, digital twin construction, and virtual tourism. Specifically, our panoramic-based partitioning strategy reduces the computational resources required for city-scale reconstruction, promoting more accessible high-quality 3D content creation. While the technology could theoretically be used for unauthorized surveillance, the risks are consistent with existing 3D reconstruction methods and can be mitigated by standard data privacy protocols, such as blurring faces and license plates in the input panoramic images.

\subsection{Limitations}
Despite its performance, \textit{PanoLOG} has several limitations:
\begin{itemize}
    \item \textbf{Dynamic Objects:} Like most 3DGS-based methods, our framework assumes a static scene. Transient objects (e.g., moving vehicles or pedestrians) may cause ``ghosting'' artifacts if not pre-processed with masks.
    
    \item \textbf{Memory Constraints:} Although {\boldmath$G^2$}\textbf{PS} enables block-wise training, the global coarse training stage in Stage~I still requires a certain amount of GPU memory for very large-scale initializations.
\end{itemize}

\label{supp:impact}
% 必填项：讨论社会影响与局限性

%\newpage
%\input{checklist.tex}

\end{document}